\documentclass[letterpaper]{article} 
\usepackage{aaai2026}  
\usepackage{times}  
\usepackage{helvet}  
\usepackage{courier}  
\usepackage[hyphens]{url}  
\usepackage{graphicx} 
\usepackage{natbib}  
\usepackage{caption} 
\usepackage{algorithm}
\usepackage{algorithmic}

\usepackage{microtype}
\usepackage{booktabs}
\usepackage{tabularx}  
\usepackage{amsmath}  
\usepackage{proof}    
\usepackage{multirow}
\usepackage{subcaption}
\usepackage{tcolorbox}
\usepackage{array}
\usepackage{enumitem}
\usepackage{amssymb}
\usepackage{dashrule} 

\usepackage{newfloat}
\usepackage{listings}
\DeclareCaptionStyle{ruled}{labelfont=normalfont,labelsep=colon,strut=off} 
\floatstyle{ruled}
\newfloat{listing}{tb}{lst}{}
\floatname{listing}{Listing}
\title{LexInstructEval: Lexical Instruction Following Evaluation for \\ Large Language Models}
\author{
    Huimin Ren\textsuperscript{\rm 1}\equalcontrib,
    Yan Liang\textsuperscript{\rm 2}\equalcontrib\\
    Baiqiao Su\textsuperscript{\rm 2},
    Chaobo Sun\textsuperscript{\rm 1}\thanks{Chaobo Sun is the corresponding author.},
    Hengtong Lu\textsuperscript{\rm 1},
    Kaike Zhang\textsuperscript{\rm 1},
    Chen Wei\textsuperscript{\rm 1}
}
\affiliations{
    \textsuperscript{\rm 1}Li Auto Inc.,\\
    \textsuperscript{\rm 2}Beijing University of Posts and Telecommunications\\
    {\{renhuimin, luhengtong, zhangkaike, chenwei10\}@lixiang.com}, \\
    {\{liangyan523, subaiqiao\}@bupt.edu.cn},
    {sunchaobo@gmail.com}

}

\usepackage{bibentry}

\begin{document}

\maketitle

\begin{abstract}
The ability of Large Language Models (LLMs) to precisely follow complex and fine-grained lexical instructions is a cornerstone of their utility and controllability. However, evaluating this capability remains a significant challenge. Current methods either rely on subjective and costly human evaluation or on automated ``LLM-as-a-judge'' systems, which suffer from inherent biases and unreliability. Existing programmatic benchmarks, while objective, often lack the expressiveness to test intricate, compositional constraints at a granular level. To address these limitations, we introduce LexInstructEval, a new benchmark and evaluation framework for fine-grained lexical instruction following. Our framework is built upon a formal, rule-based grammar that deconstructs complex instructions into a canonical (Procedure, Relation, Value) triplet. This grammar enables the systematic generation of a diverse dataset through a multi-stage, human-in-the-loop pipeline and facilitates objective verification via a transparent, programmatic engine. We release our dataset and open-source evaluation tools to facilitate further research into the controllability and reliability of LLMs.
\end{abstract}
\begin{links}
    \link{Code}{https://github.com/huiminren/LexInstructEval}
\end{links}

\section{Introduction}

\begin{figure}[t!] 
\centering
\includegraphics[width=0.95\linewidth]{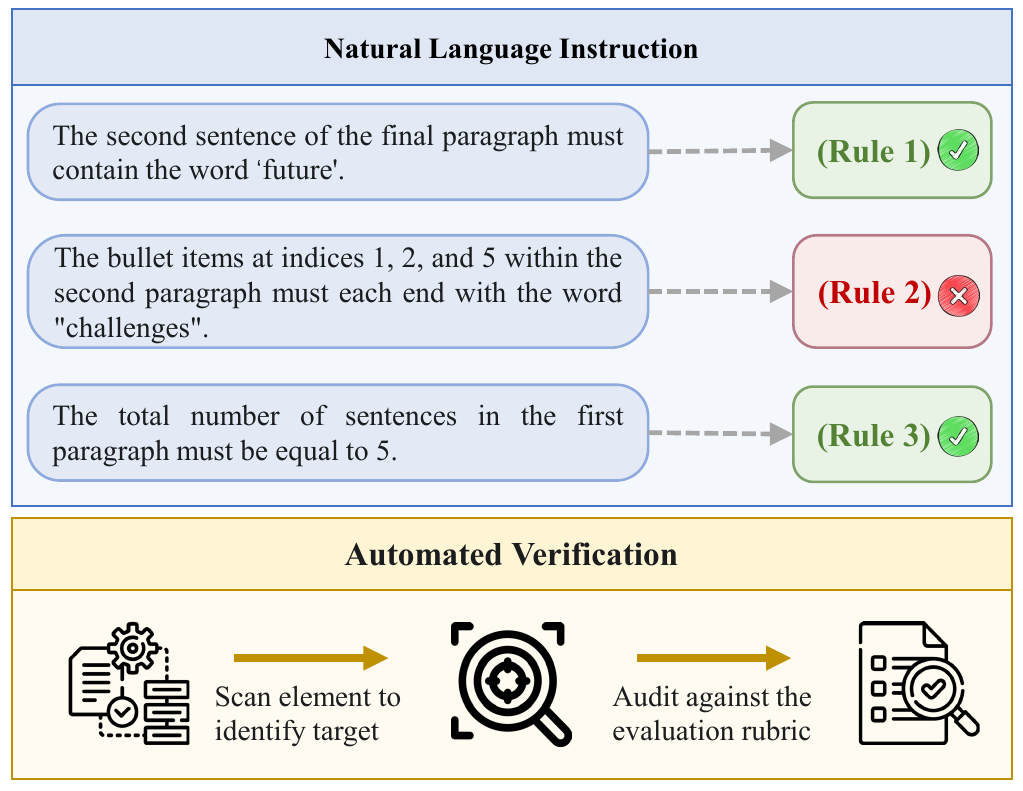} 
\caption{A motivating example illustrating the core idea of LexInstructEval. Complex natural language instructions are decomposed into fine-grained, verifiable rules, which are then checked by our automated verification engine.}
\label{fig:introduction}
\end{figure}

The advent of Large Language Models (LLMs) has been propelled by their remarkable ability to follow human instructions. This capability, cultivated through techniques like instruction tuning \cite{wei2021finetuned} and Reinforcement Learning from Human Feedback (RLHF) \cite{ouyang2022training}, has transformed LLMs from mere text completers into versatile AI assistants. The precision with which a model can adhere to complex and specific user directives is a direct measure of its utility and controllability. Consequently, the rigorous evaluation of this instruction-following ability has become a paramount challenge in the field.

However, existing evaluation methodologies exhibit significant limitations when faced with fine-grained, verifiable lexical constraints. Human evaluation, while considered the gold standard, is notoriously slow, expensive, and subject to inter-annotator variability, making it impractical for large-scale model development \cite{chang2024survey}. In response, the community has largely pivoted to automated evaluation. Yet, this has introduced a new set of challenges. The ``LLM-as-a-judge'' paradigm, employed by prominent benchmarks like FollowBench \cite{jiang2023followbench}, offers scalability but suffers from inherent biases and has been shown to be unreliable for verifying hard, objective constraints \cite{zheng2023judging}. On the other hand, existing programmatic benchmarks like IFEval \cite{zhou2023instruction}, while objective, often lack the granular expressiveness required to test complex, compositional instructions that target specific textual units. This leaves a critical gap: the absence of a benchmark that can precisely and objectively measure an LLM's ability to follow intricate lexical commands at multiple levels of granularity.

To bridge this gap, we introduce \textbf{LexInstructEval}, a novel benchmark framework designed to systematically evaluate an LLM's ability to follow fine-grained lexical instructions. As illustrated in Figure~\ref{fig:introduction}, our approach conceptualizes complex instruction following as a series of verifiable sub-tasks. At the core of our framework is a formal, rule-based grammar that deconstructs complex natural language instructions into a canonical, machine-readable triplet: $\langle \texttt{Procedure, Relation, Value} \rangle$. This structure allows us to represent a wide array of constraints—from the number of paragraphs down to the content of a single character—with unambiguous precision. Evaluation is conducted by a transparent and efficient verification engine. Unlike resource-intensive LLM-based judges, our purely rule-based engine is extremely low-cost and fast. To validate its objectivity, we conducted a study showing that its verdicts achieve \textbf{97\% consistency} with expert human annotators, confirming its reliability without compromising on efficiency.

Our main contributions, which directly address the limitations of prior work, are threefold:
\begin{itemize}[leftmargin=*, topsep=3pt, itemsep=2pt]
    \item \textbf{A Formal Grammar for Lexical Constraints:} We propose a novel, expressive grammar that can systematically represent complex, compositional instructions, surpassing the granularity of existing programmatic benchmarks. This serves as the foundation for both generating and verifying intricate lexical tasks.

    \item \textbf{A High-Quality, Human-Validated Benchmark:} We construct and release LexInstructEval, a new benchmark developed through a systematic pipeline that prevents logical conflicts. Crucially, every instruction is validated by human experts to ensure clarity and fairness, a step often overlooked in fully automated data synthesis.

    \item \textbf{An Efficient and Reliable Verification Engine:} We provide our open-source evaluation tool, built on an architecture designed for rubric-based adjudication that involves element isolation and target identification. In contrast to subjective or resource-heavy methods, our engine is highly efficient and demonstrably reliable, achieving 97\% agreement with human judgment.
\end{itemize}

\section{Related Work}

\subsubsection{Evaluation of LLMs.}The evaluation of Large Language Models (LLMs) is a multifaceted challenge, broadly categorized into two primary approaches: human evaluation and automatic evaluation. Human evaluation is widely regarded the gold standard, as it can capture nuanced aspects such as creativity, coherence, and alignment with human values, which are difficult to quantify automatically. For example, the development of models like InstructGPT \cite{ouyang2022training} relied heavily on human preference data to align the behavior of the model. However, this approach is inherently slow, costly, and difficult to scale. To overcome these limitations, automatic evaluation methods are employed. The initial methods adapted traditional NLP metrics such as BLEU \cite{papineni2002bleu}, but these are often insufficient for open tasks. Consequently, the field has gravitated towards two main automatic approaches: standardized benchmarks like MMLU \cite{hendrycks2020measuring} for knowledge assessment and the emergent ``LLM-as-a-judge" paradigm \cite{zheng2023judging}. The latter uses a powerful proprietary model as a proxy for human judgment, offering a scalable and cost-effective method that correlates surprisingly well with human preferences \cite{gu2024survey}.

\subsubsection{Instruction-Following Evaluation.}Evaluating an LLM's adherence to instructions requires specialized benchmarks. Recent efforts have made significant strides, IFEval employs verifiable, programmatic checks to test for hard constraints \cite{zhou2023instruction}, while FollowBench \cite{jiang2023followbench} and research on multi-constraint composition \cite{wen2024benchmarking} primarily use powerful LLMs as automated judges. While these methods provide crucial scalability, they introduce other limitations. LLM-as-a-judge systems can be unreliable and exhibit significant biases, such as a preference for longer outputs, which may not correlate with factual correctness or true instruction adherence \cite{zheng2023judging}. To overcome the limitations of existing automated evaluators, our work introduces a new benchmark built on a meticulously designed, rule-based framework. It features instructions with finer granularity precise to the sentence level and is assessed via controllable rules. This approach is low-cost, fast, and objective. The rules are human-validated to ensure high consistency, and the benchmark’s architecture is easily extensible to multiple languages.

\section{Approach}\label{sec:approach}
\begin{figure*}[ht]
    \centering
    \includegraphics[width=\linewidth]{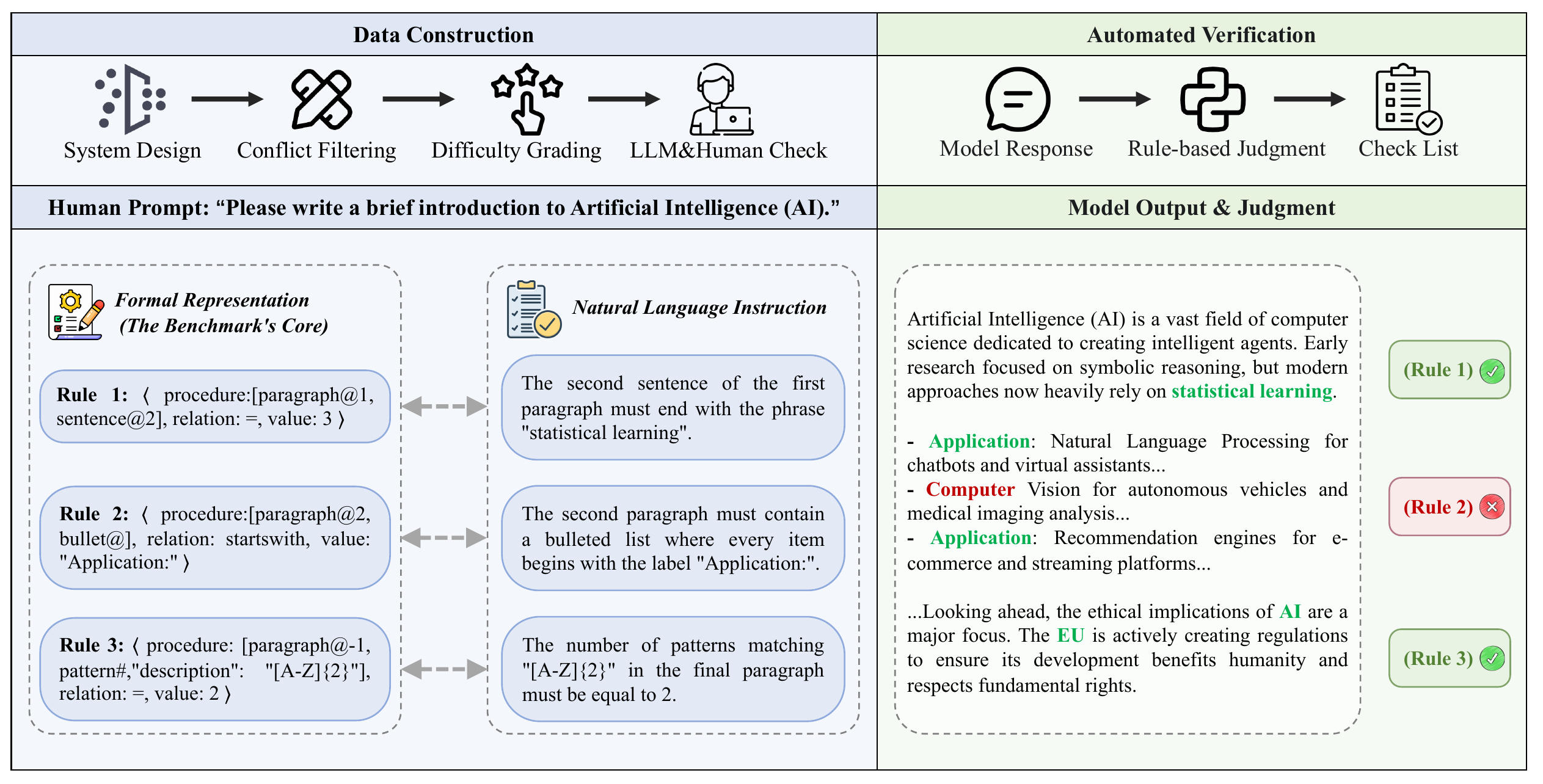} 
    \caption{The overall workflow of LexInstructEval, encompassing the four-stage data construction pipeline and the three-stage automated verification engine.}
    \label{fig:overview}
\end{figure*}

In this section, we detail the architecture of LexInstructEval. Our framework encompasses two synergistic components, as illustrated in Figure~\ref{fig:overview}: (1) a rigorous, multi-stage methodology for \textbf{Data Construction} to generate diverse and high-quality instructions, and (2) a transparent \textbf{Automated Verification Engine} to programmatically and objectively assess model adherence.

\subsection{Methodology for Data Construction}
The construction of our benchmark, LexInstructEval, follows the four-stage methodology illustrated in the lower panel of Figure.~\ref{fig:overview}. This process, designed to ensure the diversity, logical consistency and scaling of complexity of the data set, unfolds in distinct stages: (1) System design, where we define the formal grammar of the instructions; (2) Conflict filtering, which programmatically eliminates illogical or ambiguous constraints; (3) Difficulty grading, where we quantify the complexity of each instruction; and (4) LLM \& human check, a final semi-automated curation step to guarantee quality.

\subsubsection{System Design}
The foundation of our data construction is a systematic design centered on a formal grammar that deconstructs complex lexical instructions into a machine-readable, structured format. Each instruction is fundamentally represented as one or more canonical triplets: $\langle \texttt{Procedure, Relation, Value} \rangle$. This structure unambiguously defines what to check, how to check it, and what to check against.

\textbf{Procedure} A hierarchical path to locate the target text, composed of a sequence of \texttt{level} and \texttt{predicate} pairs. The \texttt{level} (Table \ref{tab:levels}) specifies the text granularity (e.g., \texttt{paragraph}, \texttt{sentence}, \texttt{word}), while the \texttt{predicate} (Table \ref{tab:predicates}) defines the selection or aggregation method (e.g., selecting the N-th element, counting occurrences).

\begin{table}[h!]
\centering

\begin{tabularx}{\linewidth}{@{} l X @{}}
\toprule
\textbf{Level} & \textbf{Description} \\ \midrule
\texttt{answer} & The entire generated response. \\
\texttt{paragraph} & A block of text separated by at least two newline characters. \\
\texttt{line} & A string of text separated by a single newline character. \\
\texttt{bullet} & An item in a Markdown list (e.g., starting with \texttt{`*`}, \texttt{`-`}, \texttt{`1.`}). \\
\texttt{sentence} & A linguistic sentence, tokenized based on language-specific rules. \\
\texttt{word} & A word, typically whitespace-delimited (for English). \\
\texttt{character} & A single Chinese character (for Chinese). \\
\texttt{letter} & A single alphabet character (for English). \\
\texttt{punc} & A single punctuation mark. \\
\texttt{pattern} & A custom pattern defined by a regular expression. \\ \bottomrule
\end{tabularx}
\caption{Hierarchy of Textual Levels}
\label{tab:levels}
\end{table}

\begin{table}[h!]
\centering

\begin{tabularx}{\linewidth}{@{} l l X @{}}
\toprule
\textbf{Predicate} & \textbf{Syntax} & \textbf{Description} \\ \midrule
Indexing & \texttt{@N}, \texttt{@} & Selects the N-th element(s). \texttt{@-1} for the last. \texttt{@} for all. \\
Before & \texttt{!N} & Selects all content before the N-th element. \\
After & \texttt{\$N} & Selects all content after the N-th element. \\
Between & \texttt{\%} & Selects the content between each element at the current level. \\
Counting & \texttt{\#} & Counts the number of elements or occurrences. \\ \bottomrule
\end{tabularx}
\caption{Predicate Definitions}
\label{tab:predicates}
\end{table}

\textbf{Relation and Value} The \texttt{relation} specifies the comparison operator, while the \texttt{value} provides the target literal for comparison.

\subsubsection{Conflict Filtering}
To ensure every generated instruction is logically sound and verifiable, we implement an automated conflict filtering stage. This stage integrates a set of validation rules directly into the generation process, preventing the formation of conflicting or ill-defined constraints.

\textbf{Type Safety Enforcement.} A key aspect of this filtering is the enforcement of type safety. We divide the relations into two mutually exclusive types. \textbf{Numerical Relations} (\texttt{>}, \texttt{<}, \texttt{=}, etc.) can only be paired with the \texttt{\#} (counting) predicate. This enforces a critical distinction: numerical operators apply only to quantitative outputs (i.e., counts), preventing illogical operations like asking if the text of a sentence is greater than 5. Conversely, \textbf{Character-based Relations} (\texttt{contain}, \texttt{startswith}, etc.) are restricted to predicates that extract textual content.

\textbf{Uniqueness and Verifiability Rules.} The filtering process further refines the applicability of character-based relations to ensure each constraint is unambiguous and has a single, canonical representation (see Table \ref{tab:relations}). For instance, \texttt{startswith} and \texttt{endswith} are restricted to the \texttt{@} predicate because applying them to a block selector like \texttt{!} (before) would be redundant and programmatically ill-defined (e.g., ``content before sentence 5 starts with 'X'" is functionally identical to the simpler ``sentence 1 starts with 'X'"). These rules guarantee that every instruction is not only verifiable but also expressed in its most direct and logical form.

\begin{table}[h!]
\centering

\begin{tabularx}{\linewidth}{@{} l X @{}}
\toprule
\textbf{Predicate(s)} & \textbf{Allowed Relations} \\ \midrule
\texttt{@} & \texttt{startswith, endswith, equal, contain, notstartswith, notendswith, notcontain} \\
\texttt{!} / \texttt{\$} & \texttt{contain, notcontain, equal} (\texttt{equal} is for \texttt{\$} only) \\
\texttt{\%} & \texttt{equal} \\ \bottomrule
\end{tabularx}
\caption{Applicability of Character-based Relations to Predicates}
\label{tab:relations}
\end{table}

\subsubsection{Difficulty Grading}
To enable a fine-grained analysis of model capabilities, each instruction is automatically assigned a difficulty level (\texttt{Easy}, \texttt{Medium}, or \texttt{Hard}) through a comprehensive, rule-based scoring system. The system quantifies complexity by evaluating each constraint across several key dimensions: the structural depth of its \texttt{Procedure}, the cognitive load of its \texttt{Predicate} (e.g., interval selection \texttt{\%} vs. simple indexing \texttt{@N}), the strictness of its \texttt{Relation} (e.g., equal vs. contain), and the complexity of its \texttt{Value} (e.g., regex vs. a single character). The total difficulty score is an aggregate of these factors, further amplified by a multiplier that increases with the number of constraints in an instruction to account for inter-constraint dependencies. 

\subsubsection{LLM \& Human Check}
The final stage is a crucial semi-automated quality check, where machine readable rules are translated into human readable prompts and then rigorously reviewed for fluency, clarity, and fairness.

The process begins by programmatically generating a vast set of seed instructions in the structured $\langle \texttt{Procedure, Relation, Value} \rangle$ format, with initial ideas sourced from the Infinity-Instruct dataset~\cite{li2025infinityinstructscalinginstruction}. To convert these machine-readable rules into prompts for the LLMs, we employ a set of \textbf{curated natural language templates}. Each template maps a specific combination of grammatical components to a fluent sentence structure. For example, a rule triplet for counting sentences is rendered into text like, ``The response must contain exactly 5 sentences.'' This template-based approach ensures consistency and clarity across the entire dataset.

Once translated into natural language, the prompts undergo a dual-check process. First, an initial automated check is performed by Qwen3-235B-A22B~\cite{qwen3} to filter for awkward phrasing or obvious semantic contradictions. Following this LLM pre-screening, the refined instructions undergo a final, thorough manual review by expert annotators. This human-in-the-loop curation is critical for two reasons. First, it filters out subtle logical inconsistencies or ambiguous wording. Second, annotators ensure each instruction is \textbf{explicit and self-contained}, detailing all requirements to eliminate ambiguity as a potential cause for non-compliance (see Figure~\ref{fig:overview} for examples). This comprehensive process guarantees that every instruction in LexInstructEval is challenging, fair, and clearly articulated.

\subsection{Final Dataset Composition}
\begin{table}[h!]
\centering

\begin{tabular}{l ccc}
\toprule
\textbf{Difficulty Level} & \textbf{English} & \textbf{Chinese} & \textbf{Total} \\ \midrule
Easy & 321 & 332 & 653 \\
Medium & 372 & 372 & 744 \\
Hard & 550 & 528 & 1,078 \\ \midrule
\textbf{Total} & \textbf{1,243} & \textbf{1,232} & \textbf{2,475} \\
\bottomrule
\end{tabular}
\caption{Distribution of instructions in the LexInstructEval benchmark by difficulty level and language.}
\label{tab:dataset_distribution}
\end{table}

This rigorous process culminates in the final LexInstructEval benchmark, which comprises two parallel, monolingual datasets: one in English and one in Chinese. Importantly, these are \textbf{independently constructed datasets}, not translations of each other, to ensure linguistic and cultural authenticity. The distribution of instructions across difficulty levels is presented in Table~\ref{tab:dataset_distribution}. The intentional skew towards the Hard category ensures our benchmark can effectively test and differentiate state-of-the-art models.

\subsection{Automated Verification Engine}
To operationalize our formal grammar, we designed an automated verification engine that translates each structured rule into a definitive boolean judgment. The engine is designed for maximum transparency, efficiency, and objectivity. To validate its reliability, we conducted a study comparing its verdicts against those of human experts on a sample of 500 instances, achieving a 97\% agreement rate. This confirms that our automated approach serves as a reliable proxy for human judgment.

The engine's architecture conceptualizes verification as a three-stage logical pipeline: \textbf{Element Isolation}, \textbf{Target Transformation}, and \textbf{Rubric Adjudication}.

\subsubsection{Element Isolation via Procedural Parsing}
The initial stage aims to precisely isolate the relevant text segment(s) from the full model response, guided by the rule's \texttt{Procedure} path. This is not a single-step split, but an iterative context refinement. The engine begins with the entire response as its initial context. It then sequentially executes each operation in the \texttt{Procedure} array, progressively drilling down through the textual hierarchy (answer $\rightarrow$ paragraph$\rightarrow$ sentence, etc.) to narrow the analytical scope.

\subsubsection{Target Identification within Isolated Elements}


Once the relevant elements are isolated, this stage transforms them into the final entity for evaluation. The most significant operation is the counting predicate (\texttt{\#}), which triggers a text-to-value transformation by computing a numerical value. If no such quantitative operator is present, the target is the textual content itself. The outcome is the final subject of adjudication: either an integer or a string.

\subsubsection{Adjudication against the Evaluation Rubric}
In the final stage, the engine performs the ultimate validation by applying the rule's relation and value to the target identified in the previous step. The adjudication logic is type-driven, bifurcating its approach based on the target's data type, thereby enforcing the type safety established in our grammar.

\begin{itemize}[leftmargin=*, topsep=3pt, itemsep=2pt]
    \item \textbf{Numerical Adjudication:} If the target is an integer (from a count), a direct mathematical comparison(e.g., \texttt{=} , \texttt{>}, \texttt{<=})is performed against the rule's value.

    \item \textbf{Textual Adjudication: }If the target is text, a string-based comparison (e.g., \texttt{startswith}, \texttt{contain}, \texttt{equal}) is applied.
\end{itemize}

A core tenet of this stage is the principle of universal quantification. When a \texttt{Procedure} is meant to apply to a set of elements (e.g., all paragraphs), the adjudication succeeds only if every single element in the set satisfies the constraint. A failure in one element results in the failure of the entire rule, mirroring the logical $\mathtt{\forall}$ (for all) quantifier.

To formalize this three-stage logic, we present the core verification process as a high-level algorithm. Algorithm~\ref{alg:verification} illustrates the conceptual journey from a raw text input to a final boolean judgment, mapping directly to our Isolate-Identify-Adjudicate framework.

\begin{algorithm}[H]
\caption{Conceptual Verification Flow}
\label{alg:verification}
\begin{algorithmic}[1]
\REQUIRE \textit{rule}: verification rule; \textit{full\_text}: response text
\ENSURE  A boolean judgment
\STATE \textbf{function} Verify(\textit{rule}, \textit{full\_text})
    \STATE \textbf{Step 1: Element Isolation}
    \STATE \textit{scope} $\gets$ \textit{full\_text} 
    \FOR{each \textit{step} in \textit{rule.procedure}}
        \STATE \textit{scope} $\gets$ RefineScope(\textit{scope}, \textit{step})
    \ENDFOR

    \STATE \textbf{Step 2: Target Identification \& Adjudication}
    \IF{\textit{rule} requires counting}
        \STATE \textit{target} $\gets$ CountElements(\textit{scope}, \textit{rule})
        \STATE \textit{judgment} $\gets$ NumericalCompare(\textit{target}, \textit{rule})
    \ELSE
        \STATE \textit{target} $\gets$ GetTextContent(\textit{scope})
        \STATE \textit{judgment} $\gets$ TextualCompare(\textit{target}, \textit{rule})
    \ENDIF
    \STATE \textbf{return} \textit{judgment}
\end{algorithmic}
\end{algorithm}

\noindent
Algorithm~\ref{alg:verification} formalizes this verification logic. The initial for loop (lines 4--6) executes the Element Isolation stage, iteratively refining the scope based on the procedure. The subsequent if/else block (lines 8--13) is responsible for both Target Identification and Adjudication. It achieves this by first discerning if the target is numerical (the result of a count) or textual (the content itself) and then dispatching to the corresponding comparison logic.


\section{Experiments}\label{sec:results}

\begin{table*}[!th]
\centering
\begin{tabular}{@{} l @{\hspace{3.5cm}} ccc ccc c @{}}
\toprule
\multirow{2}{*}{\textbf{Model}} & \multicolumn{3}{c}{\textbf{Strict Acc (\%)}} & \multicolumn{3}{c}{\textbf{Loose Acc (\%)}} & \multirow{2}{*}{\textbf{\shortstack{Gain \\ (\%)}}} \\
\cmidrule(lr){2-4} \cmidrule(lr){5-7}
 & \textbf{CN} & \textbf{EN} & \textbf{Overall} & \textbf{CN} & \textbf{EN} & \textbf{Overall} & \\
\midrule
\addlinespace
\multicolumn{8}{l}{\textit{Non-Reasoning-oriented Open-source Models}} \\
\addlinespace 
Qwen2.5-7B         & 22.1 & 23.4 & 22.7 & 24.4 & 25.5 & 25.0 & 2.2 \\
Qwen2.5-14B        & 21.9 & 23.7 & 22.8 & 24.7 & 26.3 & 25.5 & 2.7 \\
Qwen2.5-32B        & 25.9 & 25.1 & 25.5 & 28.4 & 27.6 & 28.0 & 2.5 \\
Qwen2.5-72B        & 25.6 & 25.7 & 25.6 & 28.8 & 28.0 & 28.4 & 2.8 \\
kimi-k2-0711-preview & 38.0 & 34.3 & 36.1 & 40.3 & 37.2 & 38.8 & 2.6 \\
DeepSeek-V3-0324   & 30.8 & 28.9 & 29.9 & 35.4 & 34.3 & 34.8 & 4.9 \\
\midrule
\addlinespace
\multicolumn{8}{l}{\textit{Reasoning-oriented Open-source Models}} \\
\addlinespace
QwQ-32B            & 35.5 & 32.6 & 34.1 & 39.2 & 36.2 & 37.7 & 3.6 \\
Qwen3-8B           & 34.1 & 33.5 & 33.8 & 38.5 & 36.5 & 37.5 & 3.7 \\
Qwen3-14B          & 37.0 & 34.3 & 35.6 & 41.0 & 38.0 & 39.5 & 3.9 \\
Qwen3-30B-A3B      & 37.6 & 34.7 & 36.2 & 41.4 & 38.6 & 40.0 & 3.8 \\
Qwen3-32B          & 34.1 & 32.3 & 33.2 & 37.8 & 35.3 & 36.5 & 3.3 \\
Qwen3-235B-A22B    & 35.5 & 33.7 & 34.6 & 39.0 & 36.9 & 37.9 & 3.4 \\
R1-Distill-Qwen-7B & 20.1 & 17.8 & 18.9 & 22.0 & 19.5 & 20.8 & 1.8 \\
R1-Distill-Llama-8B& 20.2 & 17.3 & 18.8 & 22.2 & 19.2 & 20.7 & 1.9 \\
R1-Distill-Qwen-14B& 22.6 & 21.4 & 22.0 & 25.4 & 23.5 & 24.4 & 2.4 \\
R1-Distill-Qwen-32B& 24.9 & 23.5 & 24.2 & 28.6 & 26.1 & 27.4 & 3.2 \\
R1-Distill-Llama-70B& 24.3 & 23.4 & 23.8 & 26.9 & 25.4 & 26.2 & 2.3 \\
R1-0528-Qwen3-8B   & 27.4 & 21.3 & 24.3 & 31.8 & 25.6 & 28.7 & 4.3 \\
DeepSeek-R1-0528   & 44.0 & 33.4 & 38.7 & 50.2 & 38.3 & 44.2 & 5.6 \\
\midrule
\addlinespace
\multicolumn{8}{l}{\textit{Closed-source Models}} \\
\addlinespace
Doubao-seed-1.6-thinking & 38.2 & 36.0 & 37.1 & 43.5 & 39.3 & 41.4 & 4.3 \\
Claude-Sonnet-3.7  & 35.6 & 29.8 & 32.7 & 42.0 & 34.8 & 38.4 & 5.7 \\
Claude-Sonnet-4    & 39.7 & 31.7 & 35.7 & 45.6 & 36.2 & 40.9 & 5.2 \\
Claude-Opus-4      & 40.4 & 33.7 & 37.0 & 45.8 & 37.4 & 41.6 & 4.5 \\
Gemini-2.5-Flash   & 50.3 & 40.2 & 45.3 & 56.1 & 45.1 & 50.6 & 5.3 \\
Gemini-2.5-Pro     & 52.0 & 49.7 & 50.9 & 61.7 & 56.5 & 59.1 & \textbf{8.2} \\
GPT-4o-2024-11-20  & 29.1 & 26.6 & 27.8 & 31.9 & 29.7 & 30.8 & 3.0 \\
GPT-4.1-2025-04-14 & 37.7 & 32.1 & 34.9 & 42.4 & 36.8 & 39.6 & 4.7 \\
GPT-o1-2024-12-17  & 71.4 & 61.7 & 66.6 & 73.9 & 65.2 & 69.6 & 3.0 \\
GPT-o3-2025-04-16  & \textbf{76.9} & \textbf{63.5} & \textbf{70.2} & \textbf{79.5} & \textbf{69.0} & \textbf{74.2} & 4.0 \\
GPT-o3-mini        & 70.0 & 59.3 & 64.6 & 73.0 & 62.6 & 67.8 & 3.1 \\
GPT-o4-mini        & 73.5 & 61.1 & 67.3 & 76.3 & 65.0 & 70.7 & 3.4 \\
\bottomrule
\end{tabular}
\caption{Benchmark results across different accuracy metrics, with each score averaged over four independent runs.}
\label{tab:main_results_final}
\end{table*}

\subsection{Experimental Setup}
\subsubsection{Evaluation Metrics}
To quantify model performance, we adopt a dual-metric approach inspired by IFEval~\cite{ifeval}, allowing for assessment at different levels of stringency. Our evaluation is centered on a programmatic verification engine that checks for adherence to given constraints. Based on this, we define two metrics:

\begin{itemize}[leftmargin=*, topsep=3pt, itemsep=2pt]
    \item \textbf{Strict Accuracy:} This metric measures the percentage of model responses that pass the verification engine without any modification. It evaluates a model's ability to follow instructions with perfect fidelity.

    \item \textbf{Loose Accuracy:} This metric offers a more lenient evaluation. It first applies a predefined set of non-semantic transformations to the model's output (e.g., stripping common markdown and conversational fillers). A response is judged as correct if any of the resulting transformed candidates passes the verification engine.
\end{itemize}

\subsubsection{Models}
We conducted a comprehensive evaluation of a wide range of models using LexInstructEval. For non-reasoning-oriented open-source models, we selected Qwen2.5-7B/14B/32B/72B\cite{Yang2024Qwen25TR}, kimi-k2-0711-preview\cite{GitHubMo43:online}, and DeepSeek-V3-0324\cite{liu2024deepseek}. For reasoning-oriented open-source models, we selected QwQ-32B\cite{team2025qwq}, Qwen3-8B/14B/32B, Qwen3-30B-A3B, Qwen3-235B-A22B\cite{yang2025qwen3}, R1-Distill-Qwen-7B/14B/32B, R1-Distill-Llama-8B/70B, R1-0528-Qwen3-8B and DeepSeek-R1-0528\cite{deepseekr1}. For closed-source models, we selected Doubao-seed-1.6-thinking\cite{bytedance2025seed}, Claude-Sonnet-3.7\cite{Claude37Sonnet}, Claude-Sonnet-4, Claude-Opus-4, Gemini-2.5-Flash, Gemini-2.5-Pro\cite{comanici2025gemini}, GPT-4o-2024-11-20\cite{hurst2024gpt}, GPT-4.1-2025-04-14\cite{achiam2023gpt}, GPT-o1-2024-12-17\cite{jaech2024openai}, GPT-o3-2025-04-16\cite{OpenAIgpto3}, GPT-o3-mini-2025-01-31 and GPT-o4-mini-2025-04-16.

\subsubsection{Inference Configurations}
Closed-source models were queried via API, while open-source models were deployed locally using the SGLang framework. We adhered to all officially recommended hyperparameters. If such recommendations were unavailable for an open-source model, we adopted the settings from closed-source models as a baseline. The maximum output length (\texttt{max\_tokens}) was configured per model category: 16,384 for reasoning models and 8,192 for general-purpose models. Full details regarding our experimental setup for reproducibility are provided in the code link.

\subsection{Results and Analysis}
\begin{figure}[ht]
\centering
\includegraphics[width=\linewidth]{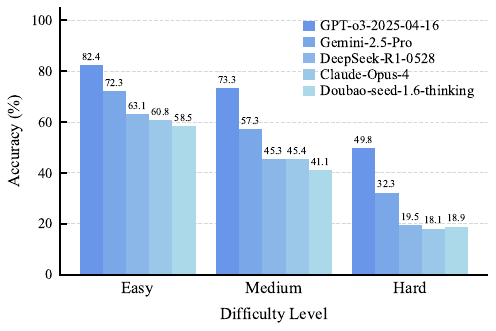}
\caption{Strict Accuracy of evaluated models across different difficulty levels.}
\label{fig:difficulty}
\end{figure}

Our analysis unfolds in a structured progression, starting from a high-level overview of model performance and then drilling down into the nuanced factors that influence instruction-following capabilities. We first dissect the aggregate data from Table~\ref{tab:main_results_final} to understand overall rankings and key model-level trends. Subsequently, we leverage our visualizations (Figures~\ref{fig:difficulty} and \ref{fig:four_subfigures}) to explore how performance is affected by the intrinsic complexity of the instructions themselves.

\subsubsection{Overall Performance} 
Table~\ref{tab:main_results_final} presents the overall performance of all models on LexInstructEval, with results averaged over \textbf{four} independent tests. GPT-o3-2025-04-16 achieves the highest overall performance, demonstrating SOTA capabilities in both Chinese (76.9\%) and English (63.5\%). Several key trends emerge from the results:

\begin{itemize}[leftmargin=*, topsep=3pt, itemsep=2pt]
    \item \textbf{A Challenging and Discriminative Benchmark:} Following complex lexical instructions remains a formidable challenge, with the top score at just 70.2\%. The wide performance range (from under 20\% to over 70\%) confirms our benchmark's ability to effectively discriminate between models of varying capabilities.
    
    \item \textbf{Significant Cross-Lingual Disparity:} A model's performance is shown to be highly language-dependent. Top models like GPT-o3 score significantly higher on Chinese than on English tasks (e.g., a 13.33-point gap). This empirical evidence demonstrates that monolingual evaluation can lead to a skewed assessment, underscoring the necessity of our bilingual benchmark.
    
    \item \textbf{Superiority of Reasoning-Oriented Models:} Models explicitly fine-tuned for reasoning consistently outperform their non-reasoning counterparts, suggesting that skills for multi-step logical planning are highly transferable to mastering compositional instructions.
    
    \item \textbf{Varying Degrees of Output Fidelity:} The `Gain` column (uplift from Strict to Loose Accuracy) reveals differences in output precision. Models like Gemini-2.5-Pro exhibit a high gain (8.2\%), suggesting they are "sloppy but smart," often failing on precise formatting. In contrast, the low gain of the GPT-o family (3-4\%) indicates higher output fidelity and control.
\end{itemize}

\subsubsection{Impact of Instruction Difficulty} 
To further investigate the factors driving this performance disparity, we analyze model accuracy across varying levels of instruction difficulty, as illustrated in Figure~\ref{fig:difficulty}. A universal trend is evident: the performance of all evaluated models systematically degrades as the difficulty increases from "Easy" to "Hard," validating our difficulty metric. More importantly, the performance gap between the top-tier models and others widens dramatically when faced with the most challenging instructions. While state-of-the-art models like GPT-o3-2025-04-16 (49.8\%) and Gemini-2.5-Pro (32.3\%) exhibit a relatively graceful degradation, the other models, including DeepSeek-R1-0528 and Claude-Opus-4, suffer a sharp performance collapse, with their accuracies plummeting to below 20\% on "Hard" tasks. This suggests that the key differentiator for leading models is their robustness and resilience when confronted with highly complex instructions.

\begin{figure}[htbp]
    \centering
    \begin{subfigure}[b]{0.48\linewidth}
        \centering
        \includegraphics[width=\linewidth]{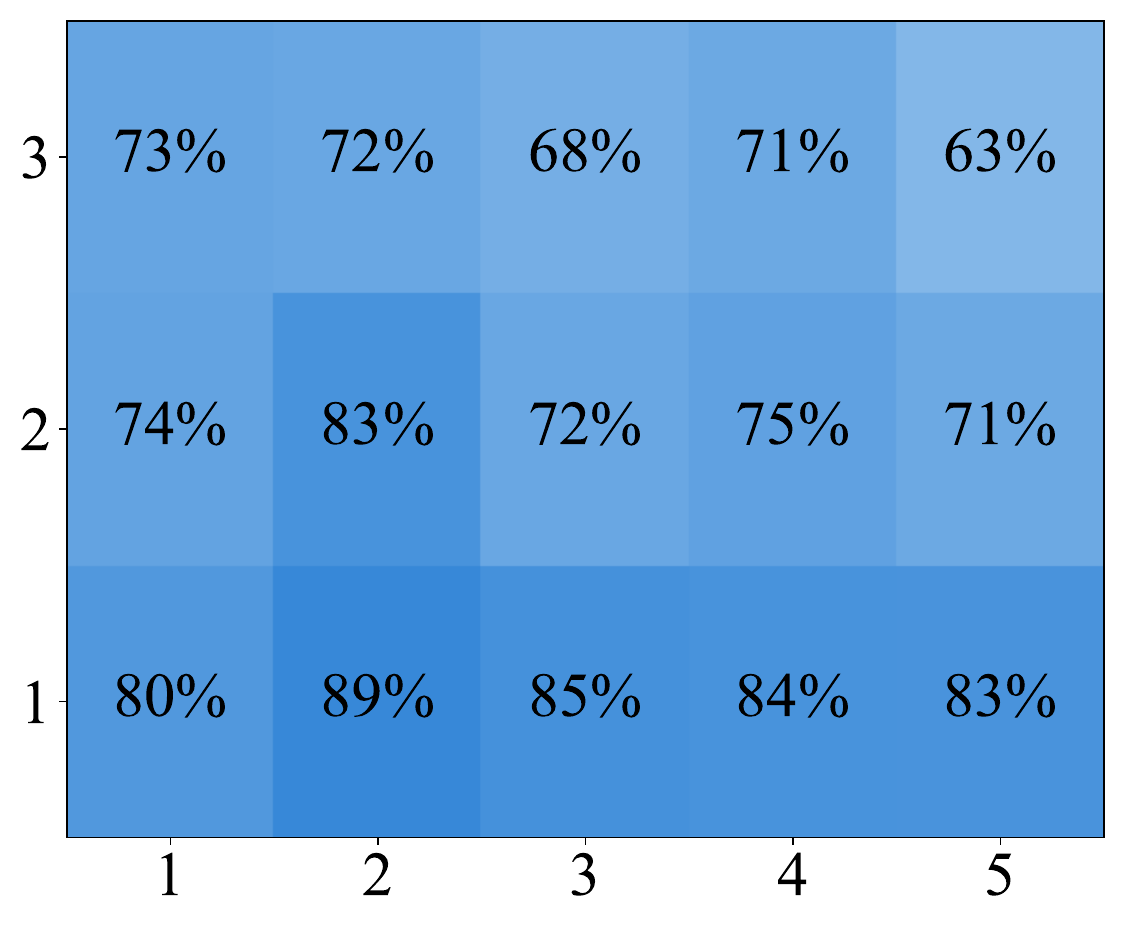}
        \caption{GPT-o3 on English Dataset}
        \label{fig:sub_gpto3_en}
    \end{subfigure}
    \hfill
    \begin{subfigure}[b]{0.48\linewidth}
        \centering
        \includegraphics[width=\linewidth]{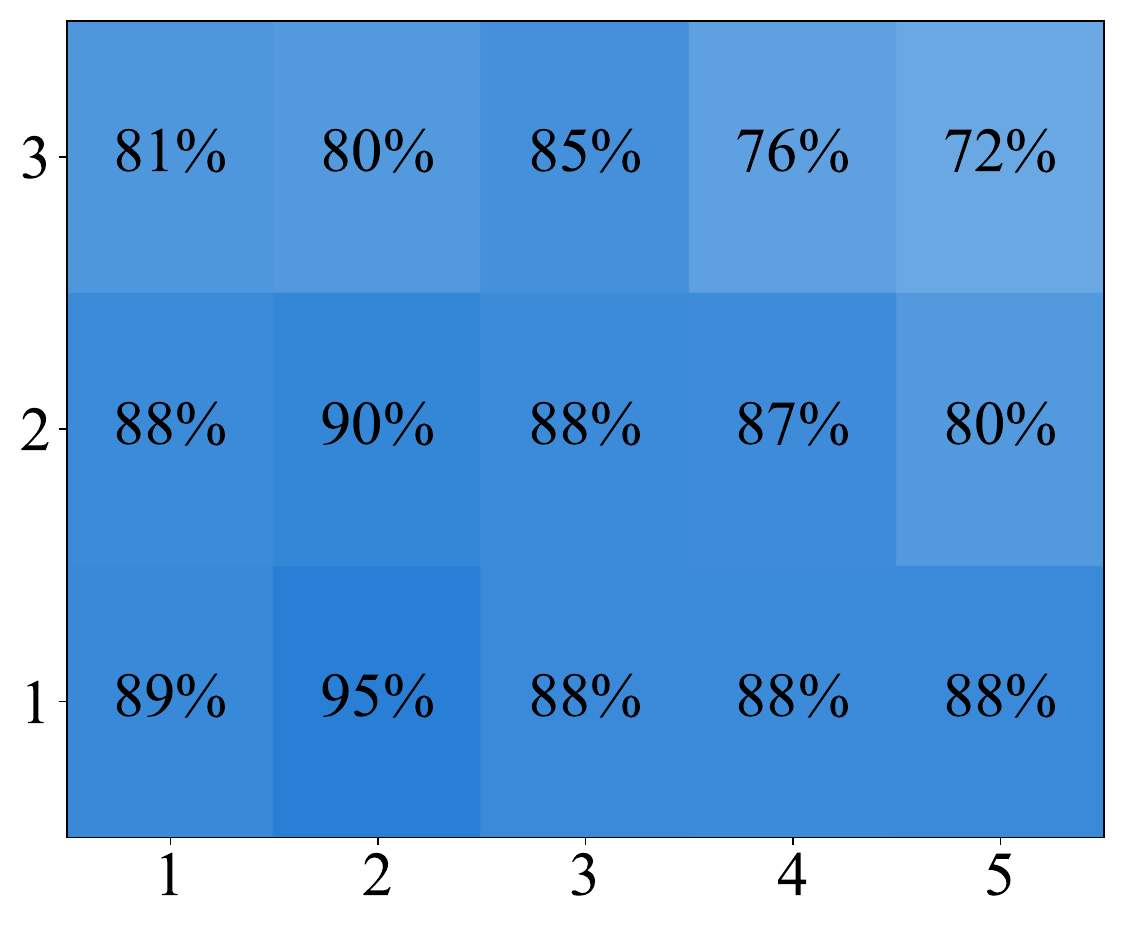}
        \caption{GPT-o3 on Chinese Dataset}
        \label{fig:sub_gpto3_zh}
    \end{subfigure}
    \vskip\baselineskip
    \begin{subfigure}[b]{0.48\linewidth}
        \centering
        \includegraphics[width=\linewidth]{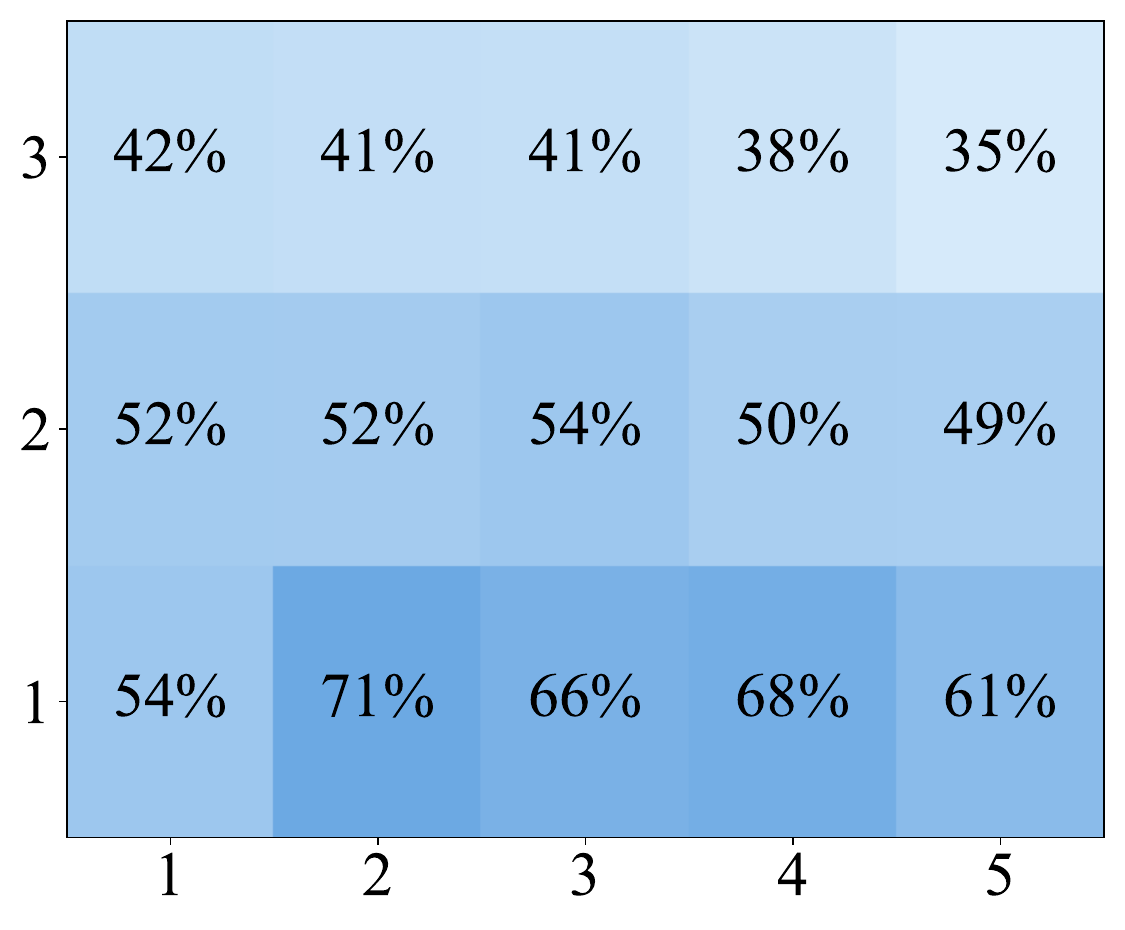}
        \caption{DS-R1 on English Dataset}
        \label{fig:sub_r1_en}
    \end{subfigure}
    \hfill
    \begin{subfigure}[b]{0.48\linewidth}
        \centering
        \includegraphics[width=\linewidth]{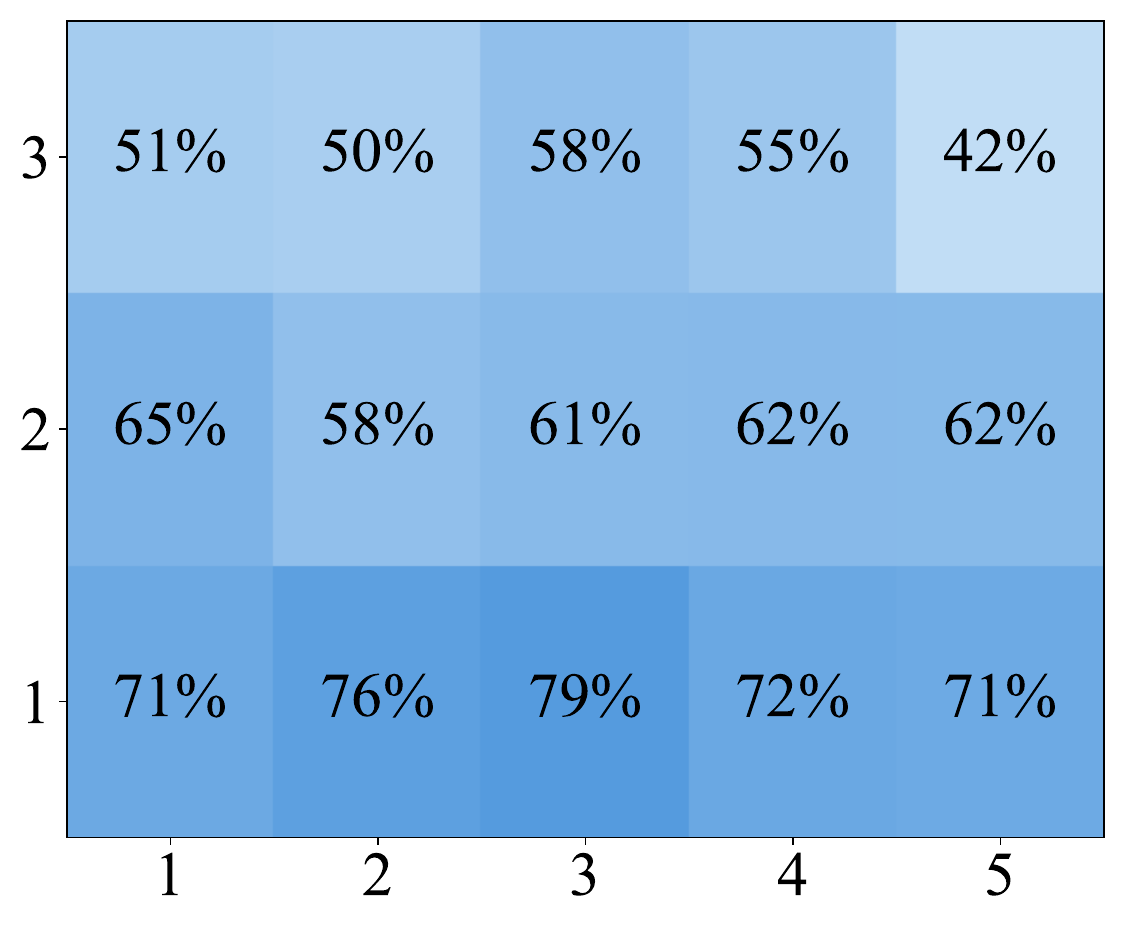}
        \caption{DS-R1 on Chinese Dataset}
        \label{fig:sub_r1_zh}
    \end{subfigure}
    \caption{Heatmap analysis of Strict Accuracy, broken down by Procedure Depth (y-axis) and Instruction Count (x-axis). }
    \label{fig:four_subfigures}
\end{figure}

\subsubsection{Performance under Compounding Complexity} 
To diagnose the specific failure points of models under increasing cognitive load, we detail their performance against both instruction breadth and depth in Figure.~\ref{fig:four_subfigures}. While all heatmaps show the expected performance decay from the bottom-left (low complexity) to the top-right (high complexity), a deeper analysis reveals a more nuanced story.

The primary finding is not just that GPT-o3 (~\ref{fig:sub_gpto3_en},~\ref{fig:sub_gpto3_zh}) is more capable than DeepSeek-R1 (~\ref{fig:sub_r1_en},~\ref{fig:sub_r1_zh}), as indicated by its substantially darker heatmaps, but how their capabilities differ. A closer look suggests that procedural depth is a more formidable challenge than instruction breadth. For both models, increasing the nesting level of a single command (moving vertically up a column) typically results in a steeper accuracy drop than merely adding more shallow instructions (moving horizontally across a row).

This exposes the key differentiator: GPT-o3 demonstrates far greater resilience to logical depth. It maintains high performance even when tasked with multiple, deeply nested instructions. In contrast, DeepSeek-R1’s performance degrades sharply as soon as procedural depth increases, indicating a lower threshold for parsing and executing complex, hierarchical commands. This suggests the core advantage of state-of-the-art models lies not just in handling more instructions, but in their superior ability to process deep logical structures. This proficiency is critical, as it directly enables a more complete understanding and execution of tasks at a finer granularity, such as manipulating a specific word within a sentence or a character within a word.

\section{Conclusion}\label{sec:conclusion}


In this paper, we addressed the critical challenge of evaluating fine-grained, compositional lexical instruction following by introducing \textbf{LexInstructEval}, a comprehensive benchmark framework. Our work delivers a three-fold contribution: (1) a novel, expressive formal grammar for representing complex instructions, (2) a high-quality, human-validated bilingual dataset, and (3) an open-source, efficient verification engine whose reliability is substantiated by a 97\% agreement rate with expert human judgment.

Our extensive evaluation using this framework reveals two key insights. First, adhering to such intricate instructions remains a \textbf{formidable challenge} even for state-of-the-art LLMs. Second, we find that the primary performance bottleneck is not the quantity of instructions but their \textbf{logical depth}. This finding highlights that true mastery of instruction following hinges on a model's ability to process deep, hierarchical commands—a capability our benchmark is uniquely designed to measure and, we hope, to help improve.

\section{Acknowledgments}
This research was fully funded and supported by Li Auto. The authors wish to thank the company for fostering an innovative environment that enabled this work.

\bibliography{aaai2026}

\end{document}